\documentclass{article}

\usepackage[dblblindworkshop,preprint]{neurips_2026}
\workshoptitle{2nd Workshop on Advances in Representation Learning for Earth Observation}
\usepackage[utf8]{inputenc} % allow utf-8 input
\usepackage[T1]{fontenc}    % use 8-bit T1 fonts
\usepackage{hyperref}       % hyperlinks
\usepackage{url}            % simple URL typesetting
\usepackage{booktabs}       % professional-quality tables
\usepackage{amsfonts}       % blackboard math symbols
\usepackage{nicefrac}       % compact symbols for 1/2, etc.
\usepackage{microtype}      % microtypography
\usepackage{xcolor}         % colors
\usepackage{float}
\usepackage{graphicx}
\usepackage[moderate]{savetrees}

\usepackage[capitalize]{cleveref}

\title{Lightweight Probabilistic Downscaling from a Deterministic Base Model}
\workshoptitle{2nd Workshop on Advances in Representation Learning for Earth Observation}

\author{%
  Joseph McLean \\
  University of Glasgow \\
  \And
  Tiffany Vlaar \\
  University of Glasgow \\
  \And
  Sigrid Passano Hellan \\
  Climate System
and Climate
Services, NORCE Research AS \\
  and Bjerknes Centre for Climate Research, Bergen, Norway \\
  \And
  Linus Ericsson \\
  University of Glasgow \\
}

\begin{document}

\maketitle

\begin{abstract}
  Climate data downscaling is the task of increasing the spatial resolution of climate data, typically by generating fine-resolution regional climate data from coarse global model output. Recent machine learning (ML) work in the related task of weather forecasting has seen significant improvements due to newly devised training methods and architectural components, but these have not yet benefited downscaling. We adapt two of these methods to create a family of lightweight probabilistic ML downscaling models built on a modified U-Net backbone and evaluate them on the CORDEX-ML-Bench suite for daily maximum temperature and precipitation across three geographic regions: the Alps, New Zealand and South Africa. We find that a two-stage training curriculum, combining deterministic pretraining with probabilistic tuning, transfers well to downscaling, beating the state-of-the-art for RMSE. Our work provides an advancement towards lightweight, probabilistic downscaling models, reducing the current trade-off between computational intensity and distributional fit.
\end{abstract}

\section{Introduction}
Climate change is one of the defining societal challenges of our time.
Making informed decisions around agricultural planning, infrastructure resilience, and preparation for extreme weather requires high-resolution regional climate data such as temperature and precipitation~\citep{rummukainen2016added}.
Yet most global climate models operate at spatial resolutions of tens to more than a hundred kilometres. Even the high-resolution experiments from the Coupled Model Intercomparison Project (CMIP), the HighResMIP experiments, are generally limited to 25--50 km~\citep{haarsma2016high}, which cannot provide the fine-grained information needed for local impact.
Downscaling addresses this gap by learning to infer actionable fine-grained regional data--commonly down to the scale of tens of kilometres or even single digits--from coarse global model outputs~\citep{rampal2024enhancing}.

Forecasting and downscaling are both inherently uncertain tasks, and learning from historical data means learning from a sparse signal where extreme events are rare by definition. Most existing ML approaches fall into one of two groups. Deterministic models produce a single best-guess output, but minimising average error causes them to produce blurry predictions that systematically underestimate extremes~\citep{rampal2025reliable,ravuri2021skilful,subich2025fixing}. Generative models instead sample from a distribution of plausible outputs, recovering fine-scale statistics more faithfully, but at substantially higher training and inference cost~\citep{addison2026machine,rampal2026cordex}. Recent work on probabilistic forecasting has shown that this trade-off is not fundamental. U-Cast~\citep{cachay2026ucast} demonstrated that a standard U-Net, first pretrained deterministically to minimise mean absolute error (MAE) and then fine-tuned probabilistically to minimise the Continuous Ranked Probability Score (CRPS) \citep{crps} using Monte Carlo (MC) Dropout \citep{MCDropout}, achieves frontier probabilistic skill at a fraction of the cost of competing generative models. MOSAIC~\citep{mosaic} further showed that learned functional perturbations and block-sparse attention address the issue of overly smooth predictions without sacrificing efficiency. In this paper, we bring these ideas to the downscaling domain where they have not yet been tested.

% Our contributions are as follows:
% \begin{itemize}
%     \item We adapt the two-stage deterministic-to-probabilistic training curriculum from forecasting to downscaling, introducing a family of lightweight probabilistic downscaling models. 
%     \item We compare the effect of instantiating the resulting framework with different stochastic mechanisms (MC dropout and learned functional perturbations) 
%     and attention variants (standard and block-sparse).
%     \item  We find MC dropout with block sparse attention sets a new state of the art on RMSE on CORDEX-ML-Bench for both temperature and precipitation target variables, while remaining lightweight enough to train on a single GPU. It remains poor on many distributional metrics, however, where it is substantially outperformed by the fully generative competitor.
% \end{itemize}

Our contributions are as follows: 1) We adapt the two-stage deterministic-to-probabilistic training curriculum from forecasting to downscaling, introducing a family of lightweight probabilistic downscaling models. 2) We compare the effect of instantiating the resulting framework with different stochastic mechanisms (MC dropout and learned functional perturbations) and attention variants (standard and block-sparse). 3) We find MC dropout with block sparse attention sets a new state-of-the-art on RMSE on CORDEX-ML-Bench for both temperature and precipitation target variables, while remaining lightweight enough to train on a single GPU. Further research is needed to advance its performance on many distributional metrics, where it is substantially outperformed by the computationally expensive, fully generative competitor.

\section{Related Work}
Deep learning has been applied to statistical downscaling since early CNN-based methods~\citep{vandal2017deepsd}, with subsequent work establishing stronger architectures including convolutional U-Nets and vision transformers for a range of surface variables~\citep{banomedina2022deepesd, prasad2024evaluating}. A common approach to data-driven downscaling is to estimate the high resolution targets by minimising the mean or absolute error. This produces deterministic models with point estimates per grid point, which tend to produce blurry outputs that  underestimate extreme events~\citep{rampal2025reliable,ravuri2021skilful,subich2025fixing,Brenowitz2025}. The recent benchmark suite CORDEX-ML-Bench~\citep{rampal2026cordex} standardised comparison across this landscape, evaluating 40 ML configurations across three regions and identifying extreme-value underestimation as a consistent failure mode of deterministic models.

Generative models recover fine-scale statistics more faithfully by sampling from a distribution of plausible outputs rather than predicting a single estimate. GANs have been applied to precipitation and wind downscaling~\citep{harris2022gan,rampal2025reliable,stengel2020}, improving tail statistics at the cost of training instability. Diffusion models offer more stable training, with CorrDiff~\citep{corrdiff} demonstrating km-scale downscaling via a two-stage pipeline: a deterministic U-Net predicts the mean, and a corrective diffusion model refines the residual. While effective, these are expensive to train from scratch and not designed for lightweight deployment.

In the parallel task of weather forecasting, some recent papers propose a shift from large deterministic models~\citep{graphcast,panguweather,aurora} toward efficient probabilistic ones \citep{Brenowitz2025}. U-Cast~\citep{cachay2026ucast} showed that this does not require a dedicated generative architecture: a standard U-Net pretrained on MAE and fine-tuned on CRPS with MC dropout matches frontier probabilistic skill at a fraction of the cost. With MOSAIC, \cite{mosaic} tackle the problem of spectral degradation, the tendency of ML models to produce overly smooth predictions. Their combination of stochasticity through learned perturbations and architectural block-sparse attention matches state of the art models with an order of magnitude speed-up.

We consider a target distribution fundamentally different from that of weather forecasting. We adapt recent training and architectural advances to downscaling, where instead of predicting the next time step the model must increase the spatial resolution. This novel downscaling pipeline has the potential to significantly reduce computational cost, hence increasing the accessibility of these models.

\section{Method}
We present a simple U-Net architecture~\citep{dhariwal2021diffusion} trained via a two-stage deterministic-to-probabilistic curriculum, adapted from forecasting~\citep{cachay2026ucast}. We modify this for downscaling with consideration of additional stochastic and attention components. % to allow us to see how we can optimise our model's performance.

\paragraph{Baseline Model.}{
We adapt the U-Net used in U-Cast~\citep{cachay2026ucast}, consisting of a four-layer encoder, a bottleneck, and a four-layer decoder. All of these layers are built from foundational U-Net blocks which perform convolutions to upsample or downsample the given data and, depending on which layer they are in, may also perform attention. The most important aspect of this model, however, is not its architecture, but rather how it is trained. We use a two-stage training process to produce a lightweight and accurate model. In stage one, the model undergoes deterministic pre-training, using a MAE loss to learn strong feature representations. In the shorter stage two, it uses a CRPS loss and MC dropout to produce ensembles of predictions, allowing the model to learn forecast uncertainties.
}

\paragraph{Adapting Model for Downscaling.}{
The base model uses a U-Net architecture to extract features from the given climate data. To adapt the model from the forecasting setup to downscaling we need an increased output resolution. We therefore introduce three extra convolutional upsampling layers in the decoder, ensuring that the model can take an input grid of $16 \times 16$ values and output one that is $128 \times 128$.
%This allows the model to learn spatial relationships and features and then use what its learned to help more accurately upsample the given image.
Aside from architectural changes, the data used to train the model is also changed. Its goal changes from predicting weather at some future time step to predicting what the given coarse resolution climate data would look like at a fine-grained regional level. We use the recently introduced benchmark suite CORDEX-ML-Bench \citep{rampal2026cordex} to expose the model to several different evaluation regions and allow comparison with a range of competing state-of-the-art methods.
}

\paragraph{Additional Components.}{
We adapt two alternative methods for stochasticity and attention from MOSAIC~\citep{mosaic}. \textit{Functional perturbations} are a probabilistic mechanism in which we inject a single vector of Gaussian noise into our network via a learned gate to be added to our intermediate feature representations, allowing the network to model the distribution through samples of noise, and to learn how strongly the perturbation should affect certain parts of the network. \textit{Block sparse attention} uses compressed, selective, and local attention to allow the model to combine coarse global context, selected distant information, and local information to efficiently capture climate relationships. This allows the model to have a better context of the data around any given input point without having to compute attention for each pair of tokens.
}

\section{Experiments}

\paragraph{Setup.}{
We evaluate our family of models to see the effectiveness of adapting two-stage training from forecasting to downscaling, and how our extra architectural components affect the models' performance.
Models are trained in accordance with CORDEX-ML-Bench's training process, following the perfect empirical statistical downscaling pseudo-reality (historial information) approach. The input data given to the model is from a global climate model that has been downscaled by a regional climate model (RCM), and then coarsened. This allows the model to train against what it knows the ``perfect'' output (from the RCM) should be. Stage one deterministic training lasts for one hundred epochs and stage two probabilistic training lasts for eight, as in \cite{cachay2026ucast}. Separate models are trained for each of the three regions of data provided (New Zealand, Alps, South Africa). Once the models have then been trained we run them through an evaluation script which computes a broad range of metrics for each model, for each region, and our reported values are the means across the regions. The experiments are run across two different target meteorological variables, daily maximum near-surface temperature (tasmax), and daily accumulated precipitation (pr). % How many parameters does the model have?
Our experiments were run on a single 24 GB NVIDIA GeForce RTX 3090, with 2 CPUs and 8 GB RAM.
}

\paragraph{Results.}{
We evaluate the models on the following metrics: TXx, PSS and IAV for tasmax, and SDII, Rx1day and LHD for pr. Furthermore, both settings have RMSE, climatological mean, RALSD and inference memory (IM) usage computed. 
RMSE evaluates the average downscaling performance; climatological mean is the RMSE of the temporally averaged targets and predictions across the test set; RALSD the spatial variability (and hence the tendency to produce overly-smooth outputs); SDII evaluates wet days, and TXx and Rx1day annual extremes; IAV evaluates the interannual variability; and PSS and LHD compare the distributional fit. We refer to the CORDEX-ML-Bench paper for the details \citep{rampal2026cordex}.
We compare our models to the top generative (RCMGEM-mv-orog) and deterministic (ParamUNET/DeepESDcrps) models from CORDEX-ML-Bench, as determined according to their average overall rank.
}

\begin{table}[tbh]
  \caption{Results for models trained for temperature (tasmax). mcdropout and mcdropout-blocksparse are most accurate for RMSE, while the generative baseline RCMGEM-mv-orog leads for other metrics.}
  \label{tab-res-temp}
  \centering
  \resizebox{1.0\textwidth}{!}{%
  \begin{tabular}{lccccccc}
    \toprule
    Model & RMSE $\downarrow$ & Clim.~Mean $\downarrow$ & TXx $\downarrow$ & RALSD $\downarrow$ & PSS $\uparrow$ & IAV $\downarrow$ & IM (GB)\\
    \midrule
    % mcdropout & 1.104  & & 0.787 & 3.517 & 0.984 & 0.462 & 0.796\\
    % mcdropout-blocksparse & 1.099 & & 0.786 & 3.845 & 0.983 & 0.463 & 0.965\\
    % perturbation & 1.243  & & 1.010 & 5.082 & 0.966 & 0.451 & 0.825\\
    % perturbation-blocksparse & 1.304  & & 1.647 & 2.384 & 0.962 & 0.437 & 0.994\\
    mcdropout & \textbf{1.10}  & 0.21 & 0.79 & \phantom{0}3.52 & 0.98 & 0.46 & 0.80\\
    mcdropout-blocksparse & \textbf{1.10} & 0.22 & 0.79 & \phantom{0}3.85 & 0.98 & 0.46 & 0.97\\
    perturbation & 1.24  & 0.48 & 1.01 & \phantom{0}5.08 & 0.97 & 0.45 & 0.83\\
    perturbation-blocksparse & 1.30  & 0.38 & 1.65 & \phantom{0}2.38 & 0.96 & 0.44 & 0.99\\
    \midrule
    RCMGEM-mv-orog & 1.13 & \textbf{0.13} & \textbf{0.36} & \phantom{0}\textbf{0.05} & \textbf{0.99} & \textbf{0.13} & -\\
    ParamUNET & 1.31 & 0.44 & 1.15 & 14.60 & \textbf{0.99} & \textbf{0.13} & -\\
    \bottomrule
  \end{tabular}%
  }
\end{table}

\begin{table}[tbh]
  \caption{Results for models trained for precipitation (pr). 
  mcdropout-blocksparse is most accurate for RMSE and LHD, while the generative baseline RCMGEM-mv-orog leads for other metrics. 
  %As for temperature, our models all perform similarly, aside from RALSD, where perturbation is significantly worse than mcdropout.  
  }
  \label{tab-res-precip}
  \centering
  \resizebox{1.0\textwidth}{!}{%
  \begin{tabular}{lccccccc}
    \toprule
    Model & RMSE $\downarrow$ & Clim.~Mean $\downarrow$ & SDII $\downarrow$ & Rx1day $\downarrow$ & RALSD $\downarrow$ & LHD $\downarrow$ &  IM (GB) \\
    \midrule
    % mcdropout & 5.635  & & 7.170 & 32.385 & \phantom{0}54.967 & 0.517 & 0.796\\
    % mcdropout-blocksparse & 5.569  & & 7.124 & 31.886 & \phantom{0}50.773 & 0.505 & 0.800 \\
    % perturbation & 5.787  & & 6.771 & 42.076 & 135.132 & 0.881 & 0.994\\
    % perturbation-blocksparse & 5.795  & & 6.366 & 40.934 & 151.355 & 1.107 & 0.995 \\
    mcdropout & 5.64  & 0.56 & 7.17 & 32.39 & \phantom{0}54.97 & 0.52 & 0.80\\
    mcdropout-blocksparse & \textbf{5.57}  & 0.58 & 7.12 & 31.89 & \phantom{0}50.77 & \textbf{0.51} & 0.80 \\
    perturbation & 5.79  & 0.98 & 6.77 & 42.08 & 135.13 & 0.88 & 0.99\\
    perturbation-blocksparse & 5.80  & 1.19 & 6.37 & 40.93 & 151.36 & 1.11 & 1.00 \\
    \midrule
    RCMGEM-mv-orog & 5.98 & \textbf{0.24} & \textbf{0.87} & \textbf{14.60} & \phantom{00}\textbf{1.47} & 0.87 & -\\
    DeepESDcrps\_IFCAv2 & 6.46 & 0.28 & 0.92 & 30.80 & \phantom{0}51.40 & 2.41 & -\\
    \bottomrule
  \end{tabular}%
  }
\end{table}

As shown in \cref{tab-res-temp,tab-res-precip}, the most accurate variation of the model is the combination of MC dropout and block sparse attention. However, the baseline setup of MC dropout and dense attention performs very similarly and is more lightweight than using block sparse attention. %This distribution of results carries on into table 2, where we can see the same trend of MC dropout with block sparse performing best.
Comparing to the baseline models, we see that a) we achieve the lowest RMSE value for tasmax and pr, even across all the models in ML-CORDEX-Bench. 
In ML-CORDEX-Bench, the generative models on average perform better than the deterministic ones, which is also reflected in the values for the highest-ranked models, as reproduced in \cref{tab-res-temp,tab-res-precip}.
We find that, aside from the new-record RMSE scores, our models' performances are closer to the deterministic models. For all but two metrics, our models score within the range of the benchmark. For IAV (interannual variability; tasmax) the worst-performing model in \cite{rampal2026cordex} achieves 0.41, not far from our scores. For SDII (wet days; pr), our models are much worse than 4.79, the worst in \cite{rampal2026cordex}.

% Compare with baselines.

% Discuss RMSE being good and other metrics being bad.

\paragraph{Ablation: Two-Stage Training.}{
We investigate the impact of two-stage training on the model's performance. We take the mcdropout two-stage variant and compare it to the same model trained using only stage 1 and only stage 2.
%will see how the model performs if it were trained using only one of the two stages, with all other parts of the model remaining unchanged.
As shown in \cref{tab-res-ablation-temp}, when only trained as a deterministic model its memory usage drops, however its overall RMSE is worse. Furthermore, when only trained as a probabilistic model, memory usage increases without the reward of a smaller RMSE. Making it also worse in comparison to two-stage training. Both results show that combining both deterministic and probabilistic training results in a better overall model.
Furthermore, in a separate ablation (see Fig.~\ref{fig:appendix_fig} of the Appendix), at 10 epochs of stage-two RALSD decreases faster compared to RMSE, suggesting that a longer stage two could bring the distributional metrics down substantially.
}

\begin{table}[t]
  \caption{Ablation results on temperature (tasmax). The latter two models are trained either on only stage one or two, for the same budget of 108 epochs. We compare to the standard two-stage model.}
  \label{tab-res-ablation-temp}
  \centering
  \resizebox{1.0\textwidth}{!}{%
  \begin{tabular}{lccccccc}
    \toprule
    Model & RMSE $\downarrow$ & Clim.~Mean $\downarrow$ & TXx $\downarrow$ & RALSD $\downarrow$ & PSS $\uparrow$ & IAV $\downarrow$ & IM (GB)\\
    \midrule
    standard two-stage & 1.10  & 0.21 & 0.79 & 3.52 & 0.98 & 0.46 & 0.80\\
    108-stage-one-only & 1.39 & 0.64 & 1.12 & 69.75 & 0.96 & 0.49 & 0.67 \\
    108-stage-two-only & 1.28 & 0.52 & 0.94 & 4.91 & 0.97 & 0.51 & 0.97  \\
    \bottomrule
  \end{tabular}%
  }
\end{table}

\section{Conclusion}
We have introduced a family of cheap and lightweight downscaling models trained via a two-stage deterministic-to-probabilistic curriculum. Our results show that this pipeline produces highly performant models on some metrics--setting new state of the art on RMSE--while falling behind on those targeting specific statistical properties of the prediction distribution. While our models improve some of these scores compared to the deterministic baseline, the top generative model, though very expensive, far exceeds them. 
%Further work will include investigating the reason for worse performance on some of the metrics, particularly IAV and SDII. 
Our results are promising, and achieved without an exhaustive search for the optimal adaption of the U-Cast model to the new downscaling domain. There are likely further benefits from tuning the length of each stage, the optimiser, the learning rate scheduling, etc. In particular, we believe a longer second stage will improve the non-RMSE metrics. Our initial study shows strong potential that future work can build upon towards broader skill across the distributional metrics.

\section*{Acknowledgements}
This work was supported by the Engineering and Physical Sciences Research Council (EPSRC) via an Undergraduate Vacation Internship hosted at The University of Glasgow.
For SPH, this study is a contribution from AIGLE, funded by NORCE Holding. She acknowledges resources provided by Sigma2 through NN11122K and NS11122K.

\newpage
\bibliographystyle{plainnat} %plainnat
\bibliography{references}

\newpage

\appendix

\section{Further Results on the Two-Stage Ablation}
\begin{figure}[h]
    \centering
    \includegraphics[width=0.65\linewidth]{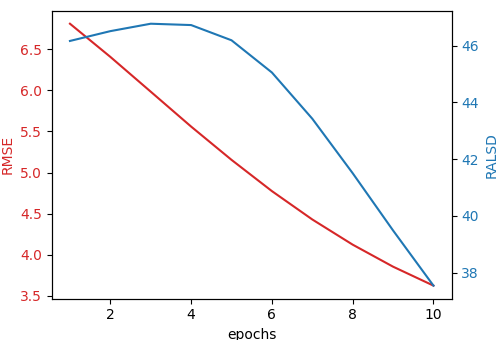}
    \caption{We track the RMSE and RALSD of our standard two-stage model during its second stage of training. The steeper downward trend for RALSD indicates that substantial improvements could be achieved by continuing training beyond the 8 epochs found to be sufficient for weather forecasting~\citep{cachay2026ucast}.}
    \label{fig:appendix_fig}
\end{figure}

To investigate the impact on the probabilistic second training stage, we plot the performance over 10 epochs of a model during the stage two training (as opposed to the original 8 in the main paper). \cref{fig:appendix_fig} shows the RMSE and RALSD metrics on separate axes for the model at every epoch. We can see that both metrics still trend downwards, but the effect is stronger for RALSD, suggesting a longer second stage could produce a model with better scores across the distributional metrics where it currently fails. Future work should examine this further, to find the optimal curriculum suited to the specific properties of the downscaling task.

%%%%%%%%%%%%%%%%%%%%%%%%%%%%%%%%%%%%%%%%%%%%%%%%%%%%%%%%%%%%

\end{document}